\documentclass[default,iicol]{sn-jnl}% Default with double column layout
    
\jyear{2022}%

\theoremstyle{thmstyleone}%
\theoremstyle{thmstyletwo}%

\theoremstyle{thmstylethree}%

\begin{document}

\title[Article Title]{Efficient Convolutional Neural Networks on Raspberry Pi for Image Classification}


\author[1]{\fnm{Rui-Yang} \sur{Ju}}\email{jryjry1094791442@gmail.com}

\author[2]{\fnm{Ting-Yu} \sur{Lin}}\email{tonylin0413@gmail.com}

\author[1]{\fnm{Jia-Hao} \sur{Jian}}\email{207440222@gms.tku.edu.tw}

\author*[1]{\fnm{Jen-Shiun} \sur{Chiang}}\email{jsken.chiang@gmail.com}

\affil[1]{\orgdiv{Department of Electrical and Computer Engineering}, \orgname{Tamkang University} \\ \orgaddress{\street{No.151, Yingzhuan Rd., Tamsui Dist.}, \city{New Taipei City}, \postcode{251301}, \country{Taiwan}}}

\affil[2]{\orgdiv{Department of Engineering Science}, \orgname{National Cheng Kung University} \\ \orgaddress{\street{No. 1 University Rd., East Dist.}, \city{Tainan City}, \postcode{70101}, \country{Taiwan}}}


\abstract{With the good performance of deep learning in the field of computer vision (CV), the convolutional neural network (CNN) architectures have become main backbones of image recognition tasks. With the widespread use of mobile devices, neural network models based on platforms with low computing power are gradually being paid attention. However, due to the limitation of computing power, deep learning algorithms are usually not available on mobile devices. This paper proposes a lightweight convolutional neural network TripleNet, which can operate easily on Raspberry Pi. Adopted from the concept of block connections in ThreshNet, the newly proposed network model compresses and accelerates the network model, reduces the amount of parameters of the network, and shortens the inference time of each image while ensuring the accuracy. Our proposed TripleNet and other State-of-the-Art (SOTA) neural networks perform image classification experiments with the CIFAR-10 and SVHN datasets on Raspberry Pi. The experimental results show that, compared with GhostNet, MobileNet, ThreshNet, EfficientNet, and HarDNet, the inference time of TripleNet per image is shortened by 15\%, 16\%, 17\%, 24\%, and 30\%, respectively. The detail codes of this work are available at \url{https://github.com/RuiyangJu/TripleNet}.}

\keywords{edge computing platform, image classification, convolutional neural network, model acceleration, model compression, Raspberry Pi.}



\maketitle

\section{Introduction}
With the rapid development of information technology, artificial intelligence has reached milestone achievements in recent years. However, the huge amount of digital data imposes a burden on computing and power consumption, reducing the model size (the number of parameters or weights of the model), is a hot research topic to improve computation efficiency and energy efficiency. Generally, an edge device needs a cloud server to complete the calculation of the model. Its disadvantage is that the data transmission cost between the edge device and the cloud server is relatively high.

In recent years, CV applications based on Raspberry Pi have begun to attract people’s attention. Bechtel \cite{bechtel2018deeppicar} simulated an autonomous-driving car by using the webcam and Raspberry Pi 3. He performed semantic segmentation tasks on Raspberry Pi to test the performance of neural networks. Monitoring systems are critical to the safety of human life, and Khalifa \cite{khalifa2019survey} compared different monitoring systems applied on Raspberry Pi and evaluated their performance. On this basis, Khalifa \cite{khalifa2021real} proposed a new CNN model to complete the human detection task on Raspberry Pi. Unmanned aerial vehicles (UAVs) can replace ordinary aircraft for search and rescue applications. The target detection task on the drone can be simulated on Raspberry Pi. Mesvan \cite{mesvan2021cnn} tested the Single Shot Detector (SSD) model on Raspberry Pi Model B, proving that their drone has an optimal detection distance of 1 to 20 meters. Caballero \cite{caballero2021inference} used a CNN model on a Raspberry Pi to classify wastes into categories such as ``plastic bottles'', ``aluminum cans'', and ``paper and cardboard'' to infer recyclable objects. However, mobile phones, mobile devices, and robotic systems are unable to use deep learning algorithms due to limited computing power. Luo \cite{luo2019cloud} combined deep learning-based CV algorithms with Raspberry Pi by utilizing the computing power of cloud servers. Although this method implements the deep neural network running on the Raspberry Pi, the data transferring between the Raspberry Pi and the cloud server consumes a lot of power. Therefore, this work aims to compress and accelerate the neural network model, and proposes a new Raspberry Pi based backbone network to promote various CV applications on Raspberry Pi.

Model compression and model acceleration are hot topics in deep neural network research. Model compression can reduce the number of parameters of neural networks, while model acceleration can reduce the inference time of neural networks. The performance of neural network models can be improved by model compression and acceleration. In order to implement vision applications in mobile devices and embedded systems, Google proposed MobileNet \cite{howard2017mobilenets}, which replaces the standard convolution layers in VGGNet \cite{simonyan2014very} with depthwise separable convolutions. A year later, Google proposed MobileNetV2 \cite{sandler2018mobilenetv2}, which introduced a linear bottleneck between the layers of the original neural network and added shortcut connections between the bottlenecks. ShuffleNet \cite{zhang2018shufflenet} uses the concept of group convolution to group feature maps to reduce computation, and therefore it can also be used in mobile phones or embedded systems. In 2019, Google once again proposed a new neural network scaling method that uses compound coefficients to uniformly scale the depth, width, and image resolution of the network. Compared with the previous SOTA network model, EfficientNet \cite{tan2019efficientnet} using this method not only improves the accuracy by 10\%, but also reduces the number of parameters by 8.4 times. In 2020, HUAWEI proposed GhostNet \cite{han2020ghostnet}, which uses the Ghost module to reduce the amount of computation. GhostNet is more accurate than MobileNetV3 \cite{howard2019searching} under similar computation. However, unlike the above methods, this work argues that reducing the connections between layers can also reduce the computation of the network model.

\begin{table*}[]
\centering
\caption{Details of TripleNet Model}
\label{tab1}
\setlength{\tabcolsep}{2.5mm}{
\begin{tabular}{|c|c|c|c|c|c|}
\hline
Model & \#Params & Flops & Layers & Channel & Growth Rate \\ \hline
TripleNet-S & 9.67M & 4.17G & 6, 16, 16, 16, 2 & 128, 192, 256, 320, 720 & 32, 16, 20, 40, 160 \\ \hline
TripleNet-B & 12.63M & 4.29G & 6, 16, 16, 16, 3 & 128, 192, 256, 320, 1080 & 32, 16, 20, 40, 160 \\ \hline
\end{tabular}
}
\end{table*}

ResNet \cite{he2016deep} builds a network model with residual learning as the main architecture, deepens the depth of the network, and achieves excellent performance in tasks such as image classification, object detection, and semantic segmentation. He \emph{et al.}, the authors of ResNet, added the split-transform-merge strategy to ResNet after referring to GoogLeNet \cite{szegedy2015going}. The newly proposed ResNeXt \cite{xie2017aggregated} has better performance with the same parameters and computation as ResNet. Zagoruyko \emph{et al.} thought from a different direction, abandoned the narrow characteristics of the ResNet model, and proposed Wide-ResNet \cite{zagoruyko2016wide}, which reduces the number of parameters and shortens the training time. DenseNet \cite{huang2017densely} passes the output of each layer to each subsequent layer to facilitate feature reuse. However, the characteristic of interconnection among all layers results in larger requirements of memory usage and overlong inference time in practical applications. In order to overcome the drawbacks of massive hardware requirement in DenseNet, HarDNet \cite{chao2019hardnet} was proposed and can reduce the connections between layers, reduce the number of data transfers, and reset the weights of layers to increase feature extraction and maintain model accuracy. ThreshNet \cite{ju2022threshnet} uses a threshold mechanism to determine the connection between layers, and resets the number of channels to obtain higher accuracy and faster inference time than that of HarDNet. However, both HarDNet and ThreshNet ignore the problems of excessively large parameters, and their applications on low-computing power platforms are limited. Based on the ThreshNet model architecture, TripleNet is proposed in this paper. This work reduces the number of parameters by improving the convolution layers in the blocks, and reduces the inference time per 100 images while improving the accuracy of the network model.

The contributions of TripleNet proposed in this paper are as follows:

\begin{itemize}
    \item [1)]
    This work proposes a new method for model compression and model acceleration. Three different convolution layers are set in different blocks, and different convolution layers use different connection methods. This variety of network architecture design reduces the computation of the network model, which is suitable for application on the platform of Raspberry Pi.
\end{itemize}

\begin{itemize}
    \item [2)]
    Different from using a cloud server to establish data transmission, this paper emphasizes the direct use of a suitable lightweight neural network on Raspberry Pi, which is more feasible with the diversified applications of embedded systems.
\end{itemize}

\section{Related Work}
Model compression is an important field for neural network research, and many research works have rarefied DenseNet through different methods. LogDenseNet \cite{hu2017log} performs sparse connections, reduces the number of input channels from $L^2$ to $Llog{L}$, and increases the output channel width to recover the accuracy dropping from the connection pruning. SparseNet \cite{liu2018sparsenet} utilizes the same sparse method as LogDenseNet, except that there is a fixed block output outputting $L+1$ layers for $L$ layers. Both models require increased growth rates to maintain accuracy, without taking the issues of computing performance and power consumption into consideration, however this sparse method is worth referring to. ConDenseNet \cite{huang2018condensenet} introduces group operation in 1 × 1 convolution and prunes the weights at the beginning of training. In addition, ConDenseNet improves on DenseNet by performing dense connection of convolutional layers across blocks. Nevertheless, it replaces the convolution layers in DenseNet, which is desirable. PeleeNet \cite{wang2018pelee} is a lightweight network variant based on Densenet, mainly for mobile devices. It proposes Stem Block to realize the downsampling of input images and the increase of channel number. This block provides stronger feature representation with less computational complexity.

The sparse connection method proposed by HarDNet \cite{chao2019hardnet} replaces the original dense connection with a harmonic dense connection scheme, and improves the output weight of the layer without sparse connection. Harmonic dense connection reduces the memory usage of the model and power consumption of hardware computing. Roy \cite{miles2021compressing} proposed the Convolution-Depthwise-Pointwise (CDP) layer, a new means of interpolating using depthwise separable convolution, and applied this method to HarDNet to reduce parameters. ThreshNet \cite{ju2022threshnet} adopts the threshold mechanism by combining dense connection and harmonic dense connection, and resets the number of channels to improve the model accuracy. It proves that the combination of dense connection and harmonic dense connection has better performance.

\begin{figure*}[h]
  \centering
  \includegraphics[width=0.95\linewidth]{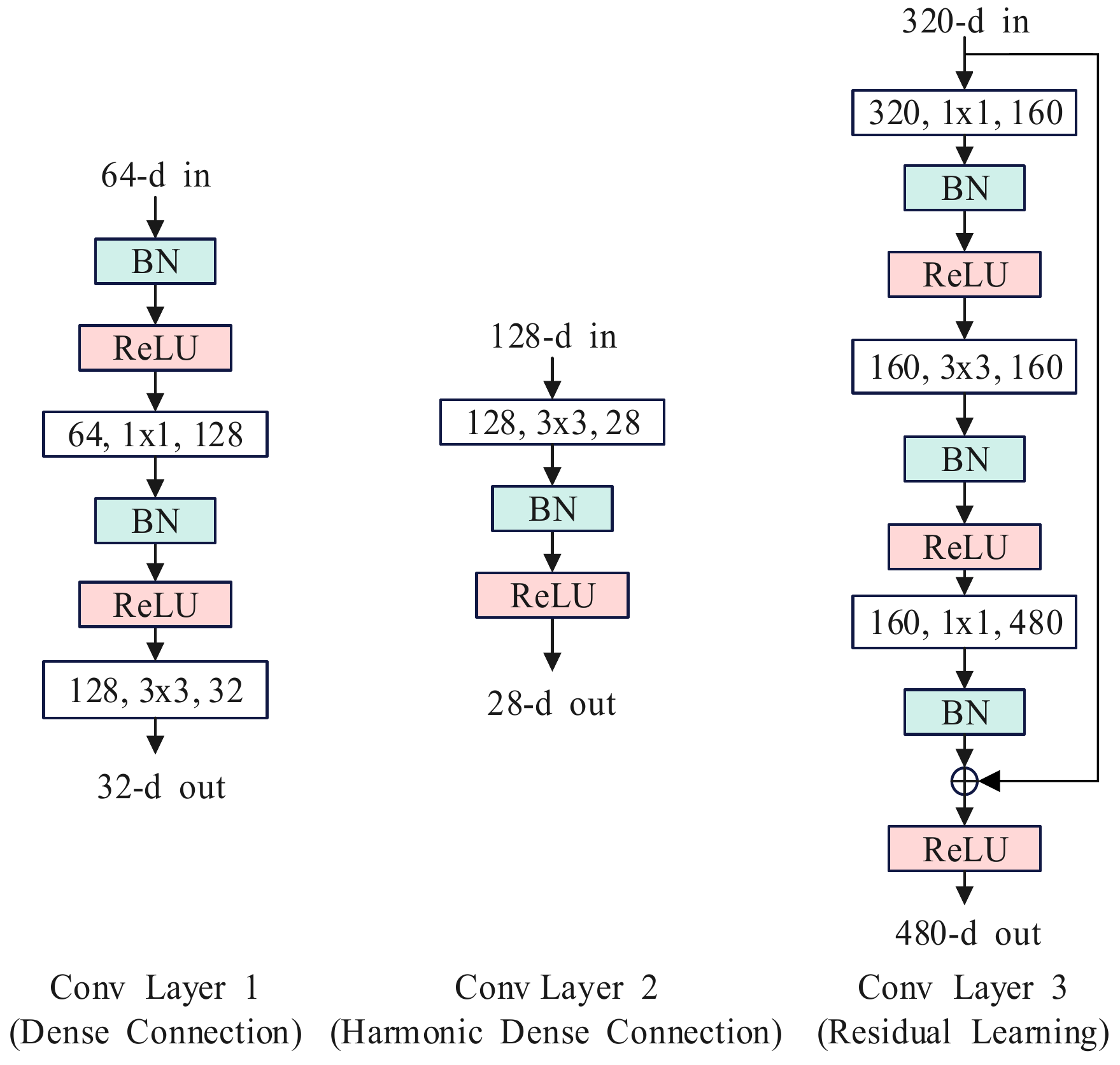}
  \caption{Three convolution layers are used by TripleNet: (a) Conv Layer 1, using dense connections, (b) Conv Layer 2, using harmonic dense connections, (c) Conv Layer 3, using residual learning.}
  \label{fig_1}
\end{figure*}

\section{Proposed Method}
\subsection{Conv Layers}
\subsubsection{Conv Layer 1}
In order to connect all the layers with the same feature map size in the block, we fix the 3 × 3 convolution output feature map as the growth rate, and the specific value of the growth rate is shown in Table \ref{tab1}. To reduce the computational complexity, we use a 1 × 1 convolution, and the output feature of the same fixed convolution is 4 × growth rate. As shown in Fig. \ref{fig_1}, BN \cite{ioffe2015batch} and ReLU \cite{glorot2011deep} are added before two convolutions, respectively. Conv Layer 1 adopts the dense connection method, and the input of the next convolution layer is the feature map of all layers:

\begin{equation}
x_n=\ H_n\left(\left[x_1,\ \cdots,\ x_{n-1}\right]\right)
\label{eq1}
\end{equation}

\begin{algorithm}
\caption{Conv Layer 2}
\label{algo}
\begin{algorithmic}[1]
\Require {Stack 3 × 3 Conv Layer in each block}
\Ensure {$3 \times 3 $ Conv - BN - ReLU}
\For{layer $n$ is even}
    \For{$i$ in $1$ to $5$}
    \State{$x=2^i=1$}
    \State{$y=n-x \ (x \leq n$)}
    \State{layer $n$ connect to layer $y$}
    \EndFor
\EndFor
\For{layer $n$ is odd}
    \State{layer $n$ connect to layer $y$}
\EndFor
\end{algorithmic}
\end{algorithm}

\begin{figure*}[h]
  \centering
  \includegraphics[width=\linewidth]{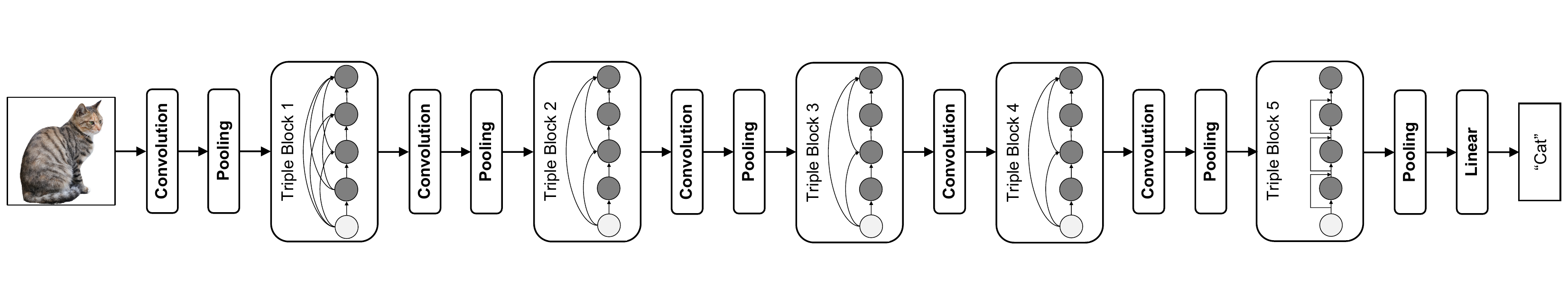}
  \caption{The overall architecture of TripleNet.}
  \label{fig_2}
\end{figure*}

\subsubsection{Conv Layer 2}
To reduce model memory usage and inference time, the convolution layers within the block are connected conditionally. Unlike Conv Layer 1, Conv Layer 2 adopts Conv-BN-ReLU, and only uses a single 3 × 3 convolution. Conv Layer 2 adopts the improved harmonic dense connection, and the fixed output of convolution with reserved input connection is 1.7 × growth rate, as shown in Fig. \ref{fig_1}, and the fixed output of convolution without input connection is the growth rate:

\begin{equation}
x_{n}=\left\{\begin{array}{c}
H_{n}\left(\left[x_{n-1}\right]\right), \text { if } n \% 2=1 \\
H_{n}\left(\left[x_{0}, \cdots, x_{n-2{ }^{i}}\right]\right), \text { if } n \% 2=0
\end{array}\right.
\label{eq2}
\end{equation}
where the even-numbered layers are connected to the previous layers in the way of the Equation \ref{eq2}, while the odd-numbered layers are only connected to the previous layer. The specific connection method is shown in Algorithm \ref{algo}.

\subsubsection{Conv Layer 3}
In order to reduce the number of parameters of the model and reduce the calculation time, the block depth is set to only 2 or 3. At the same time, to ensure the accuracy of the model, Conv Layer 3 is composed of three convolutions and uses residual learning. Since the 1 × 1 convolution is less computationally intensive, this architecture does not generate a large number of parameters. The first 1 × 1 convolution shrinks the input features by half, and the second 1 × 1 convolution triples the input features. The architecture is shown in Fig. \ref{fig_1}:

\begin{equation}
x_{n+1}=\ x_n + \mathcal{F}\left(x_n,w_n\right)
\label{eq3}
\end{equation}

\subsection{Architecture}
TripleNet consists of 5 blocks (each block is called a Triple-Block), and we propose three different convolution layers to construct these 5 blocks. After several experiments \cite{huang2017densely,ju2022threshnet,hu2017log}, Conv Layer 1 is the most effective approach to improve the accuracy of the model, but it takes up a lot of model memory, and therefore we only employ it in Triple-Block 1. Conv Layer 2 has a smaller memory usage, so we use it in Triple-Block 2, Triple-Block 3, and Triple-Block 4, which makes TripleNet retain its characteristics of smaller memory usage and suitable for edge devices. Conv Layer 3 uses residual learning, which has a smaller number of parameters. Because the input feature map of the last Triple-Block (Triple-Block 5) is large, using this convolution layer can ensure that TripleNet has a small number of parameters and a small amount of model calculation. By the mentioned scheme, the convolution layers used by the 5 blocks are determined, and the specific architecture is shown in Fig. \ref{fig_2}.

The important characteristic of CNN architecture is to obtain a smaller size feature map by downsampling to extract the features for calculation; TripleNet applies the same thought as CNN. The 1 × 1 convolution and 2 × 2 average pooling layers before the block play the role of down-sampling, and we call the combination of the convolution and average pooling layers a transition layer.

\begin{table*}[]
\centering
\caption{Detailed Implementation Parameters}
\label{tab2}
\setlength{\tabcolsep}{7mm}{
\begin{tabular}{cccc}
\toprule
Layers & Output Size & TripleNet-S & TripleNet-B \\
\midrule
\multirow{3}{*}{First Layer} & $112^2$ & \multicolumn{2}{c}{3 × 3 Convolution} \\
 & $112^2$ & \multicolumn{2}{c}{3 × 3 Convolution} \\
 & $56^2$ & \multicolumn{2}{c}{3 × 3 MaxPool} \\
\midrule 
Triple Block & $56^2$ & {[}Conv Layer 1{]} × 6 & {[}Conv Layer 1{]} × 6 \\
\midrule 
\multirow{2}{*}{Transition Layer} & $56^2$ & \multicolumn{2}{c}{1 × 1 Convolution} \\
 & $28^2$ & \multicolumn{2}{c}{2 × 2 MaxPool} \\
\midrule 
Triple Block 2 & $28^2$ & {[}Conv Layer 2{]} × 16 & {[}Conv Layer 2{]} × 16 \\
\midrule 
\multirow{2}{*}{Transition Layer} & $28^2$ & \multicolumn{2}{c}{1 × 1 Convolution} \\
 & $14^2$ & \multicolumn{2}{c}{2 × 2 MaxPool} \\
\midrule 
Triple Block 3 & $14^2$ & {[}Conv Layer 2{]} × 16 & {[}Conv Layer 2{]} × 16 \\
\midrule 
Transition Layer & $14^2$ & \multicolumn{2}{c}{1 × 1 Convolution} \\
\midrule 
Triple Block 4 & $14^2$ & {[}Conv Layer 2{]} × 16 & {[}Conv Layer 2{]} × 16 \\
\midrule 
\multirow{2}{*}{Transition Layer} & $14^2$ & \multicolumn{2}{c}{1 × 1 Convolution} \\
 & $7^2$ & \multicolumn{2}{c}{2 × 2 MaxPool} \\
\midrule 
Triple Block 5 & $7^2$ & {[}Conv Layer 3{]} × 2 & {[}Conv Layer 3{]} × 3 \\
\midrule 
\multirow{2}{*}{Classification} & $1^2$ & \multicolumn{2}{c}{AvgPool} \\
 &  & \multicolumn{2}{c}{Linear} \\
\bottomrule  
\end{tabular}
}
\end{table*}

\subsection{Detailed Design}
After extensive experiments, we propose two TripleNets, TripleNet-S and TripleNet-B, respectively, and the model details are shown in Table \ref{tab1}. TripleNet-S, ThreshNet-79 and HarDNet-68 are models of the similar order; TripleNet-B is of a higher model order. As we can see from Table \ref{tab1}, the difference between the two versions of the model is mainly in the last Triple-Block. The depth of the last Triple-Block of TripleNet-S is 2, while that of TripleNet-B is 3. As the depth increases, the numbers of channels, parameters, and the accuracy of the model all increase accordingly. Taking an image of size 224 × 224 in ImageNet as input, the feature map size extracted by TripleNet is shown in Table \ref{tab2}. A transition layer is added to the middle of each Triple-Block to obtain a smaller feature map of the input image, and the size of the feature map is $112^2$, $56^2$, $28^2$, $14^2$ and $7^2$ respectively.

\begin{table}[]
\centering
\caption{The details of the dataset}
\label{tab3}
\begin{tabular}{|c|c|c|c|c|}
\hline
Dataset & Pixel & Training set & Test set & Classes \\ \hline
CIFAR-10 & 32×32 & 50,000 & 10,000 & 10 \\ \hline
SVHN & 32×32 & 73,257 & 26,032 & 10 \\ \hline
\end{tabular}
\end{table}

\section{Experiment}
\subsection{Environment Setup}
\subsubsection{Dataset}
\begin{figure}[h]
  \centering
  \includegraphics[width=\linewidth]{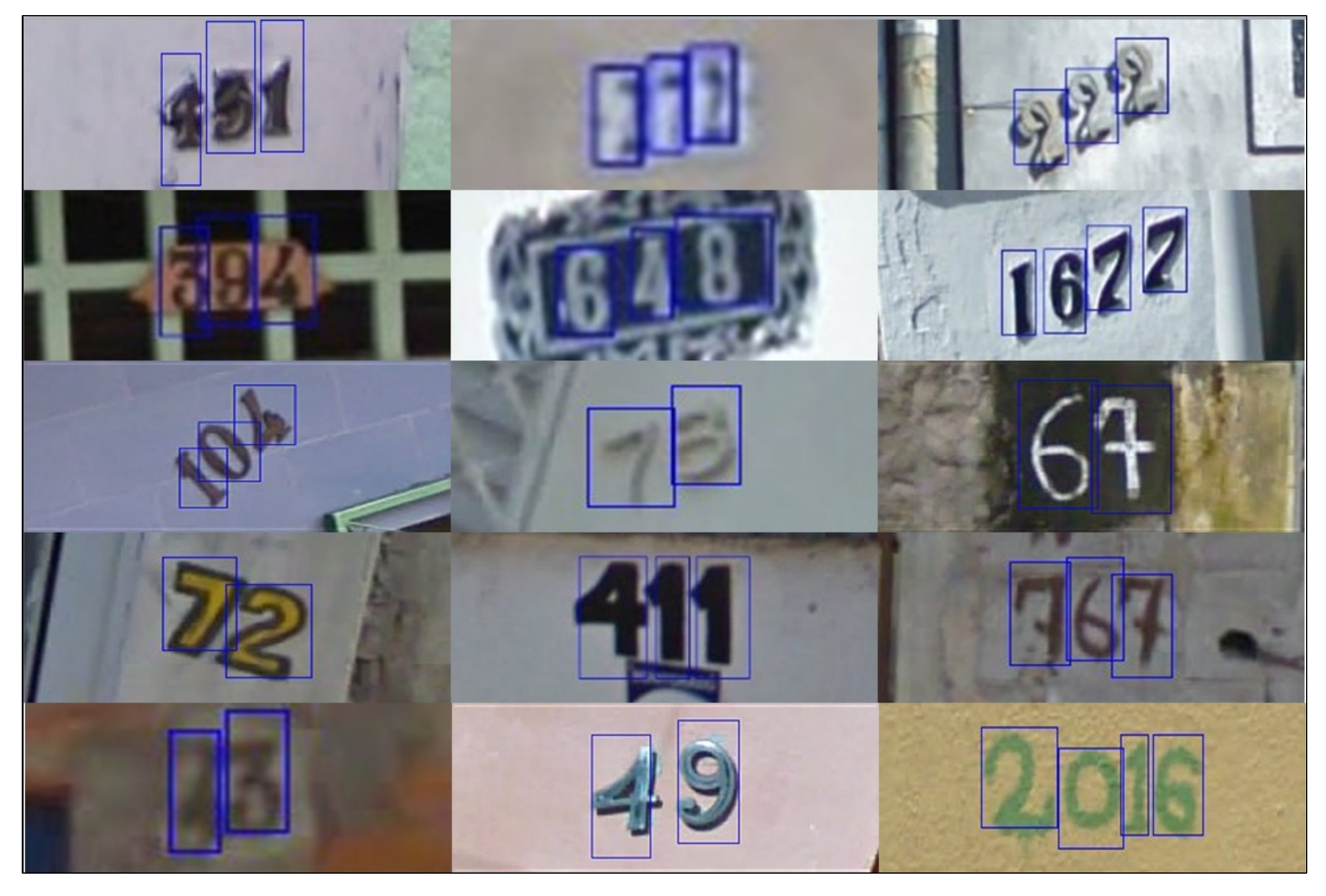}
  \caption{Examples of images in the SVHN dataset.}
  \label{fig_3}
\end{figure}

We list the training set, test set and classes of the datasets used in Table \ref{tab3}. CIFAR-10 \cite{krizhevsky2009learning} is a dataset consisting of color images of size 32 × 32, with 50,000 images for training and 10,000 images for testing, with a total of 10 classes. SVHN (Street View House Numbers) \cite{netzer2011reading} is also a dataset consisting of color images of size 32 × 32, with 73,257 training images and 26,032 testing images. As shown in Figure \ref{fig_3}, the image consists of some numbers, which are cropped from the house numbers on the street.

\begin{table*}[]
\centering
\caption{CIFAR-10 Classification Results and Model Architecture Parameters}
\label{tab4}
\setlength{\tabcolsep}{2mm}{
\begin{tabular}{|c|c|c|c|c|c|c|c|c|}
\hline
 & \begin{tabular}[c]{@{}c@{}}Inference\\ Time\\ (s)\end{tabular} & \begin{tabular}[c]{@{}c@{}}Error\\ (\%)\end{tabular} & \begin{tabular}[c]{@{}c@{}}Pre-Error\\ (\%)\end{tabular} & \begin{tabular}[c]{@{}c@{}}Flops\\ (G)\end{tabular} & \begin{tabular}[c]{@{}c@{}}MAdd\\ (G)\end{tabular} & \begin{tabular}[c]{@{}c@{}}Memory\\ (MB)\end{tabular} & \begin{tabular}[c]{@{}c@{}}\#Params\\ (M)\end{tabular} & \begin{tabular}[c]{@{}c@{}}MemR+W\\ (MB)\end{tabular} \\ \hline
TripleNet-S & 4.06 & 13.05 & 12.75 & 4.17 & 8.32 & 90.25 & 9.67 & 238.36 \\ \hline
ShuffleNet \cite{zhang2018shufflenet} & 4.41 & 13.35 & 12.71 & 2.28 & 4.55 & 83.26 & 10.18 & 221.05 \\ \hline
ThreshNet28 \cite{ju2022threshnet} & 4.53 & 14.75 & 14.49 & 2.22 & 4.31 & 617.00 & 1.01 & 1009.03 \\ \hline
TripleNet-B & 6.51 & 12.97 & 12.62 & 4.29 & 8.57 & 91.33 & 12.63 & 755.07 \\ \hline
MobileNetV2 \cite{sandler2018mobilenetv2} & 6.74 & 14.06 & 13.95 & 2.42 & 4.75 & 384.78 & 2.37 & 98.84 \\ \hline
GhostNet \cite{han2020ghostnet} & 7.67 & 19.96 & 19.75 & 0.15 & 0.29 & 40.05 & 5.18 & 251.67 \\ \hline
MobileNet \cite{howard2017mobilenets} & 7.68 & 16.12 & 15.00 & 2.34 & 4.63 & 230.84 & 3.32 & 474.13 \\ \hline
ThreshNet 95 \cite{ju2022threshnet} & 7.79 & 13.31 & 13.04 & 4.07 & 8.12 & 132.34 & 16.19 & 356.66 \\ \hline
EfficientNet B0 \cite{tan2019efficientnet} & 8.54 & 13.40 & 13.06 & 1.51 & 2.99 & 203.74 & 3.60 & 421.54 \\ \hline
HarDNet 85 \cite{chao2019hardnet} & 9.25 & 13.89 & 13.62 & 9.10 & 18.18 & 74.65 & 36.67 & 313.42 \\ \hline
\end{tabular}
}
\vspace{1mm}
\begin{tablenotes}
All network models are implemented on the same platform Raspberry Pi 4. Raspberry Pi Time is the inference time per 100 images on Raspberry Pi 4. We denote pre-training by a \textbf{Pre} mark at the beginning of the error rate. (e.g., Pre-Error).
\end{tablenotes}
\end{table*}

\subsubsection{Training}
The selection of the optimization algorithm is the top priority of the training model. Even if the same network model is trained on the same dataset, using different optimization algorithms (SGD and Adam) leads to different training results. Gradient descent is currently the most widely used optimization algorithm in neural networks. In order to make up for the shortcomings of naive gradient descent, researchers have proposed a series of variant algorithms, which have gradually evolved from Stochastic Gradient Descent (SGD) to Adam \cite{kingma2014adam}. In our experiments, all the networks are trained using Adam, which is a combination of RMSprop and Momentum. Similar to RMSprop using an exponential moving average for second-order momentum, Adam also uses an exponential moving average for first-order momentum. We set the coefficients used for computing running averages of gradient and its square to be 0.9 and 0.999, and eps (term added to the denominator to improve numerical stability) of ${10}^{-8}$. 

On CIFAR-10 and SVHN, we train models using batch size 64 for 200 and 60 epochs respectively. When training the model, the speed of gradient descent is controlled by the learning rate. The initial learning rate is set to ${10}^{-3}$, and is divided by 5 at 37.5\% and 75\% of the total number of training epochs. The loss function is used to evaluate the degree to which the predicted value of the model is different from the true value, and we use the Cross-Entropy loss function.

\subsubsection{Comparisons}
This research uses the inference time per 100 images and test error rate to evaluate our algorithm and compare with many networks, including HarDNet, ThreshNet, ShuffleNet, MobileNetV1, MobileNetV2, GhostNet, and EfficientNet.

\subsubsection{Testing}
The trained network model is evaluated for performance on Raspberry Pi 4 Model B 4GB. The Raspberry Pi is evaluated using python 3.9, torch version 1.11.0.

\begin{table*}[]
\centering
\caption{SVHN Classification Results and Model Architecture Parameters}
\label{tab5}
\setlength{\tabcolsep}{3mm}{
\begin{tabular}{|c|c|c|c|c|c|c|c|}
\hline
 & \begin{tabular}[c]{@{}c@{}}Inference\\ Time\\ (s)\end{tabular} & \begin{tabular}[c]{@{}c@{}}Error\\ (\%)\end{tabular} & \begin{tabular}[c]{@{}c@{}}Flops\\ (G)\end{tabular} & \begin{tabular}[c]{@{}c@{}}MAdd\\ (G)\end{tabular} & \begin{tabular}[c]{@{}c@{}}Memory\\ (MB)\end{tabular} & \begin{tabular}[c]{@{}c@{}}\#Params\\ (M)\end{tabular} & \begin{tabular}[c]{@{}c@{}}MemR+W\\ (MB)\end{tabular} \\ \hline
TripleNet-S & 3.87 & 6.18 & 4.17 & 8.32 & 90.25 & 9.67 & 238.36 \\ \hline
ThreshNet 28 \cite{ju2022threshnet} & 4.15 & 6.34 & 2.28 & 4.55 & 83.26 & 10.18 & 221.05 \\ \hline
ShuffleNet \cite{zhang2018shufflenet} & 4.56 & 6.34 & 2.22 & 4.31 & 617.00 & 1.01 & 1009.03 \\ \hline
MobileNetV2 \cite{sandler2018mobilenetv2} & 5.21 & 6.68 & 2.42 & 4.75 & 384.78 & 2.37 & 755.07 \\ \hline
GhostNet \cite{han2020ghostnet} & 6.62 & 8.36 & 0.15 & 0.29 & 40.05 & 5.18 & 98.84 \\ \hline
TripleNet-B & 6.62 & 5.67 & 4.29 & 8.57 & 91.33 & 12.63 & 251.67 \\ \hline
MobileNet \cite{howard2017mobilenets} & 7.38 & 7.36 & 2.34 & 4.63 & 230.84 & 3.32 & 474.13 \\ \hline
ThreshNet 95 \cite{ju2022threshnet} & 7.86 & 6.89 & 4.07 & 8.12 & 132.34 & 16.19 & 356.66 \\ \hline
EfficientNet B0 \cite{tan2019efficientnet} & 9.08 & 6.12 & 1.51 & 2.99 & 203.74 & 3.60 & 421.54 \\ \hline
HarDNet 85 \cite{chao2019hardnet} & 9.48 & 6.91 & 9.10 & 18.18 & 74.65 & 36.67 & 313.42 \\ \hline
\end{tabular}
}
\vspace{1mm}
\begin{tablenotes}
All network models are implemented on the same platform Raspberry Pi 4. Raspberry Pi Time is the inference time per 100 images on Raspberry Pi 4.
\end{tablenotes}
\end{table*}

\subsection{Experiment Results}
In Table \ref{tab4}, we show the test results of image classification on CIFAR-10 dataset. Sorted by the inference time per 100 images on Raspberry Pi, TripleNet-S has the shortest inference time of 40.6ms compared to other neural networks. Compared with ThreshNet 28, the number of parameters drops from 10.18M to 9.67M, and the error rate drops by 10\%. Although ThreshNet28 has only 4 blocks, it still generates a large amount of computation because all layers in the last block use harmonic dense connections. TripleNet-S uses 5 blocks, but the last block uses the newly proposed convolution layer, which does not generate too much computation, but the network depth is improved. We think using this approach is more reasonable than simply reducing the depth of ThreshNet. Compared with other SOTA networks, TripleNet-B has the lowest error rate of 12.97\%, and the inference time on Raspberry Pi is also lower than MobileNetV2. We think this is because the CIFAR-10 dataset has only 10 classes. If the network depth is too deep, it would lead to overfitting problems. The experimental results in Table \ref{tab4} prove that for the small dataset CIFAR-10 TripleNet is more suitable for inference on mobile devices such as Raspberry Pi. To demonstrate the above conclusions, we conduct image classification tasks on the small dataset SVHN. The test results are shown in Table \ref{tab5}. TripleNet-S performs image classification on Raspberry Pi, and the inference time per 100 images is still shorter than other neural networks, and the accuracy is not lower than ThreshNet and ShuffleNet. We emphasize that the combination of the three convolution layers reduces the number of parameters, which is an important reason for reducing inference time. The inference time of TripleNet-B is similar to GhostNet, and the error rate is lower than other SOTA neural networks. This proves that for image classification tasks, the TripleNet model performance is better than other neural networks on small datasets, and it is more suitable for running on mobile devices with limited computing power.

On CIFAR-10 dataset, we train all networks for 200 epochs, since the training loss at the 200th epoch is already close to 0.001. However, we found that DenseNet \cite{huang2017densely} used 300 epochs when training all networks on CIFAR-10. Therefore, in order to make the experiment more convincing, we pre-trained all the networks, so that the model can have a higher accuracy in training 200 epochs. As shown in Table \ref{tab4}, Pre-Error represents the test accuracy of the model with pre-training. With pre-training, the error rate of all models decreased by between 1\% and 7\%, and the model with higher testing error rate decreased more with pre-training. For example, the testing error rate of MobileNet decreased from 16.12\% to 15\% with pre-training. With pre-training, our proposed network models, TripleNet-S and TripleNet-B, drop by 2.3\% and 2.7\%, respectively. It is evident that TripleNet-B still maintains the lowest testing error rate 12.62\% after pre-training among all models.

\section{Conclusion}
As a CNN backbone network, TripleNet can be applied in different scenarios, such as the application to Mask R-CNN \cite{he2017mask} algorithm architecture to complete instance segmentation, the application to YOLOv3 \cite{redmon2018yolov3} algorithm architecture to complete object detection, the application to Panoptic FPN \cite{kirillov2019panoptic} algorithm architecture to complete the panoptic segmentation. Compared with HarDNet and ThreshNet, TripleNet has a smaller number of model parameters and can complete inference in a shorter time on Raspberry Pi. TripleNet is a network architecture designed to achieve highly efficient performance with limited computing power, which is more suitable for real-life scenarios.
As a common embedded system, Raspberry Pi can be used in many scenarios. We emphasize that inferring a neural network directly on Raspberry Pi is more efficient than using the cloud to build transmissions. The premise is that the neural network has a small amount of computation. TripleNet has efficient model performance and is the result of model compression and model acceleration on ThreshNet. TripleNet outperforms ThreshNet in terms of parameters, accuracy, and inference time.

Different from the large dataset of ImageNet, small datasets, such as CIFAR-10 and SVHN, do not require computationally expensive neural networks, which sometimes lead to poor results due to overfitting. Therefore, for image classification tasks with smaller datasets on Raspberry Pi, TripleNet is more suitable than other SOTA neural networks.

\section{Declarations}
\subsection{Funding}
The authors did not receive support from any organization for the submitted work.

\subsection{Competing interests}
The authors have no financial or proprietary interests in any material discussed in this article.

\subsection{Ethics approval}
This research does not involve human participants and/or animals.

\bibliographystyle{spmpsci}
\bibliography{reference}

\end{document}